\documentclass[10pt,letterpaper]{article}

\usepackage{ccn}
\usepackage{pslatex}
\usepackage{apacite}
\usepackage{graphicx}
\usepackage{caption}
\usepackage{blindtext}
\usepackage{amsmath}
\usepackage{tabularx}
\usepackage{fancyhdr}

\fancypagestyle{firstpagef}{%
\fancyhf{}

  \lfoot{\vspace{0.5cm}Accepted at the 2024 Conference on Cognitive Computational Neuroscience (CCN 2024)}
}

\title{Investigating the Timescales of Language Processing with EEG and Language Models}
 
\author{{\large \bf Davide Turco (davide.turco@bristol.ac.uk)}\\\large \bf Conor Houghton (conor.houghton@bristol.ac.uk)\\
Faculty of Engineering, University of Bristol, Bristol, BS8 1UB, UK}

\begin{document}

\maketitle
\thispagestyle{firstpagef}
\section{Abstract}
{
\bf
This study explores the temporal dynamics of language processing by examining the alignment between word representations from a pre-trained transformer-based language model, and EEG data. Using a Temporal Response Function (TRF) model, we investigate how neural activity corresponds to model representations across different layers, revealing insights into the interaction between artificial language models and brain responses during language comprehension.
Our analysis reveals patterns in TRFs from distinct layers, highlighting varying contributions to lexical and compositional processing. Additionally, we used linear discriminant analysis (LDA) to isolate part-of-speech (POS) representations, offering insights into their influence on neural responses and the underlying mechanisms of syntactic processing. These findings underscore EEG's utility for probing language processing dynamics with high temporal resolution. By bridging artificial language models and neural activity, this study advances our understanding of their interaction at fine timescales.
}
\begin{quote}
\small
\textbf{Keywords:} 
EEG; neurolinguistics; language models; word representations; natural language processing.
\end{quote}

\section{Introduction}
The representations of modern language models have been shown to linearly map to brain responses to the same linguistic stimulus \cite{Huth2016, Caucheteux2022}, as measured by fMRI or MEG. This may suggest that the two systems share similar mechanisms when processing language.

EEG, with its high temporal resolution, is an powerful and practical source of data for exploring the timescale of language processing in the brain. However, previous work has focused on simpler non-neural \cite{Broderick2018} or recurrent \cite{hale-etal-2018-finding} language models, often using a link function, such as surprisal \cite{turco}.

Here, we map the word representation of a pre-trained causal transformer to EEG data recorded from subjects listening to the same stimulus, via a linear convolution model. The objective of this preliminary work is to investigate the relationship between artificial models and the brain at small timescales, considering different layers of the language model and language aspects, such as syntax.

\section{Methodology}
We use publicly available EEG data \cite{bhattasali2020alice}, recorded from 52 subjects listening to the first chapter of \textit{Alice’s Adventure in Wonderland}. We limited the analysis to 33 participants, excluding subjects with excessively noisy recordings or with poor scores in the post-experiment text comprehension test. For computational purposes, the signal has been segmented into 2 s windows, with a 10\% overlap.

The same linguistic stimulus was given word-by-word to a pre-trained language model, GPT-2 \cite{radford2019language}, and word representations were extracted from the embedding layer and from a deep layer. We selected layer 8, as it has been previously shown to be one of the layers most predictive of brain responses \cite{pmlr-v139-caucheteux21a}. The activations were transformed to a vector with the same sampling frequency as the original data.

For aligning EEG activities $r_{t,e}$ and word representations $s_{t-\tau, i}$, we use a time-lagged linear regression model \cite{Crosse2016}:
\begin{equation}
    r_{t,e} = \sum_i\sum_\tau w_{\tau,e,i}\;s_{t-\tau, i}
\end{equation}
where $e$ indicates electrode and $i$ is the GPT-2 representation dimension ($i\in[0,768]$). $w$ is the linear filter kernel of length $\tau$ that, when applied to the stimulus $s$, it transforms it into the brain response $r$. This filter is known as the Temporal Response Function (TRF). 

We implemented the model in PyTorch, using a kernel ranging from -100 ms to 1 s relative to the word onset. To prevent overfitting, a L2 regularisation was applied to the model weights: the value of the regularisation coefficient was chosen among ten log-spaced from $10^{-3}$ to $10^{5}$ using 5-fold cross-validation. For each participant, a model trained with the best parameter is then tested on held-out data from that specific subject, and the Pearson correlation score between the original and reconstructed EEG signal is computed.

To isolate syntactic factors in the language model representations, we used linear discriminant analysis (LDA), motivated by previous work showing that language models encode linguistic features in a linear manner \cite{Linzen2021}. Specifically, we reduced the original 768-dimensional word representations to the same number of dimensions as the linguistic features of interest, in this case part-of-speech (POS).

\section{Results}
\begin{figure*}
    \centering
    \includegraphics[width=\linewidth]{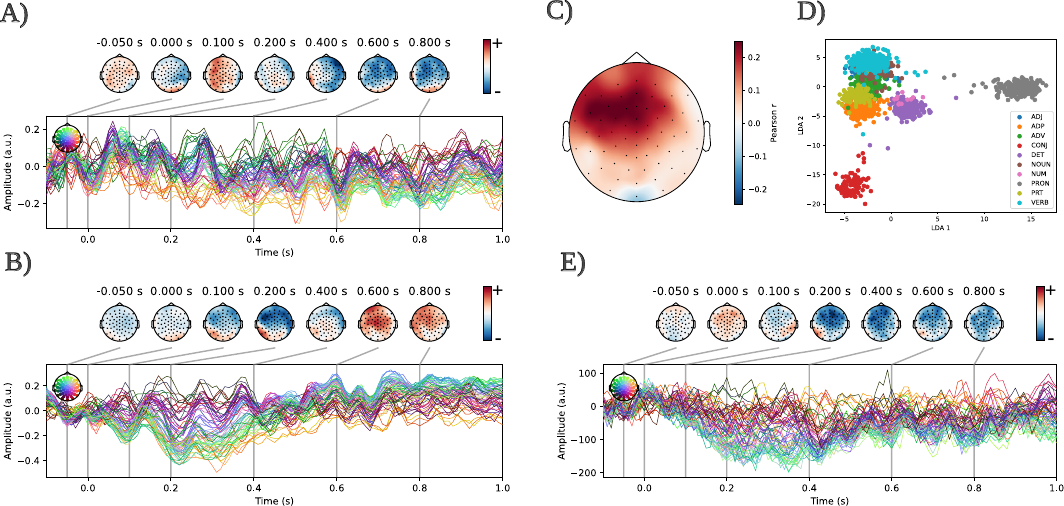}
    \caption{A) TRF obtained using the embedding layer (l=0). B) TRF obtained using a deep layer (l=8). C) Topographic map showing correlations between original and reconstructed EEG signal for a subject with high comprehension score. D) LDA-reduced representation space, with samples coloured by POS tag. E) TRF obtained using LDA-reduced representations.}
    \label{fig:results}
\end{figure*}

Our results show that mapping embedding (l=0) and deep (l=8) layer activations to EEG data lead to noticeably different TRFs. As shown in Fig. \ref{fig:results}A), the TRF for the embedding layer mostly displays negativities, especially in late time scales. On the contrary, the TRF for the deep layer (Fig. \ref{fig:results}B)) shows stronger negativity in the 200 ms region, and a positive effect in the post 600 ms region, compatible with a P600 event-related potential \cite{Coulson1998}. This shows a fundamental distinction between the embedding layer, that encodes only lexical information, and a deep layer, that encodes compositional information as well. 

In Fig. \ref{fig:results}C) we show the correlation between original and reconstructed EEG signal for a subject with high comprehension score, using activations from layer 8. The topographic plot shows higher scores in the central and left-temporal regions, normally associated with language processing. The correlation for all subjects is $0.03$ ($p\ll10^{-5}$); per-subject correlations and p-values have been aggregated using the Fisher's method.

To motivate the use of LDA to isolate syntactic representation, we plotted the reduced representation space in Fig. \ref{fig:results}D). Samples corresponding to different POS tags are tightly clustered. Interestingly, samples corresponding to content words (e.g. \texttt{NOUN}, \texttt{VERB}, \texttt{ADV}) appear closer together than short function words such as conjunctions (\texttt{CONJ}) and pronouns (\texttt{PRON}).

We then fitted our model on this reduced space and the corresponding TRF is shown in Fig. \ref{fig:results}E). The representation dimensions related to part of speech appear to negatively correlate with EEG activities in the post 200 ms segment. 

\section{Discussion}
In this paper, we have introduced an approach for investigating the timescale of language processing by mapping the word representations of a transformer-based language model to high-temporal-resolution EEG data from human participants. We have shown that embedding and deep layers lead to different responses, both in their topographic distribution and in their timescale. We have also presented a simple technique for isolating POS representations in the language model activations, and shown that these are negatively correlated with EEG activity.

In future work, we plan to improve the mapping model by adding non-linear components to the architecture. We would also like to extend the analysis to other aspects of language, like semantics. 

\section{Acknowledgements}
DT is funded by an UKRI Centre of Doctoral Training grant (EP/S022937/1). This work was carried out using the HPC facilities of the ACRC, University of Bristol. We are grateful to Dr Stewart whose philanthropy supported the purchase of GPU nodes.

\setlength{\bibleftmargin}{.125in}
\setlength{\bibindent}{-\bibleftmargin}

\bibliographystyle{apacite.bst}
\bibliography{bibliography}

\newpage
\appendix
\onecolumn
\section{Supplementary material\protect\footnote{Not included in the version of the paper accepted at CCN 2024.}}
\subsection{Glossary}
\textit{Compositionality}: the principle that the meaning of a complex expression (such as a sentence) is determined by the meanings of its constituent expressions (words) and the rules used to combine them.\\
\textit{Event-Related Potentials (ERP)}: time-locked neural signals that are averaged across multiple instances of a specific event or stimulus to identify consistent neural responses associated with that event.\\
\textit{Linear Discriminant Analysis (LDA)}: a statistical method used to find a linear combination of features that characterises or separates two or more classes.\\
\textit{Part of Speech (POS)}: annotations that indicate the grammatical category of each word in a text (e.g., noun, verb, adjective). These tags help in parsing sentences and understanding syntactic structures.\\
\textit{Semantics}:  the study of meaning in language, including how words, phrases, and sentences convey meaning.\\
\textit{Syntax}:  the set of rules, principles, and processes that govern the structure of sentences in a language, including word order and the relationships between words.\\
\textit{Temporal Response Function (TRF)}: a linear model that maps the relationship between continuous stimuli, such as language, and the corresponding neural responses over time.

\subsection{Additional methodology and implementation details}
The dataset was downloaded and processed using the openly available \texttt{Eelbrain} pipeline\footnote{\url{https://github.com/Eelbrain/Alice/tree/main}}. Additional processing of the EEG data was performed using \texttt{MNE  1.5.1}.

We obtained word representations using the GPT-2 pre-trained model, as offered by Hugging Face's \texttt{transformers 4.36.2} API. Syntactic representations were obtained by applying an LDA using \texttt{scikit-learn 1.3.2}, setting the number of components equal the number of POS tags (nine).

The TRF estimator model was implemented in \texttt{PyTorch 2.1.0} with \texttt{CUDA 12.1}, and trained on a GPU. The learning rate was set to $10^{-4}$ and the batch size to 64.

\subsection{Interpreting the TRF}
In a continuous setting, the TRF $w(t,\tau)$ is the linear mapping between a stimulus $s(t)$, in this paper a language model representation of the linguistic stimulus, and a neural response $r(t)$, as measured via e.g. EEG, for pre-determined time lags $\tau$. For instance, the TRF at $t=$ 400 ms describes the relationship between the EEG signal and the word representation vector 400 ms after, or equivalently, how linguistic stimulus at time $t$ would modulate the EEG amplitude 400 ms earlier.

The units of the amplitude of the TRF are given by the ratio of the units of the EEG data ($\mu V$) and those of the word representations. In this study, following \citeA{Broderick2018}, we standardised both the EEG signal for each channel and the word representations; the TRF is therefore presented in arbitrary units. A positive polarity, represented as red in Figs. \ref{fig:results}A, \ref{fig:results}B, \ref{fig:results}E, should be interpreted as a positive increase in EEG amplitude caused by a change in word representations of one unit.
\end{document}